\newif\ifreview
\reviewfalse

\ifreview
\documentclass[preprint,12pt]{elsarticle}
\else
\documentclass[final,1p,times]{elsarticle}
\fi

\usepackage{amsmath,amssymb,amsfonts}
\usepackage{graphicx}
\usepackage{booktabs}
\usepackage[nopatch=footnote]{microtype}
\PassOptionsToPackage{hyphens}{url}
\usepackage[hypertexnames=false]{hyperref}
\usepackage{xurl}
\usepackage{xcolor}
\usepackage{enumitem}
\usepackage{algorithm}
\usepackage{algpseudocode}
\usepackage{siunitx}
\usepackage{setspace}
\usepackage{array}
\usepackage{colortbl}
\ifreview
\usepackage{lineno}
\fi

\hypersetup{
  colorlinks=true,
  linkcolor=blue,
  citecolor=blue,
  urlcolor=blue
}


\newcommand{\R}{\mathbb{R}}
\definecolor{tableblue}{HTML}{EEF5FF}
\definecolor{tablegray}{HTML}{F7F8FA}
\newcolumntype{L}[1]{>{\raggedright\arraybackslash}p{#1}}
\newcolumntype{C}[1]{>{\centering\arraybackslash}p{#1}}
\makeatletter
\def\ps@pprintTitle{%
  \let\@oddhead\@empty
  \let\@evenhead\@empty
  \let\@oddfoot\@empty
  \let\@evenfoot\@oddfoot}
\makeatother

\begin{document}

\begin{frontmatter}

\title{Deployment-Oriented Session-wise Meta-Calibration for Landmark-Based Webcam Gaze Tracking}

\author[aff1]{Chenkai Zhang\corref{cor1}}
\cortext[cor1]{Corresponding author.}
\ead{ck.zhang26@gmail.com}
\address[aff1]{Independent Researcher, Wenzhou, Zhejiang 325000, China}

\begin{abstract}
Practical webcam gaze tracking is constrained not only by prediction error but also by calibration burden, robustness to head motion and session changes, runtime footprint, and the feasibility of browser-side deployment on commodity devices.
We therefore study a deployment-oriented operating point rather than the raw-image, large-backbone regime emphasized by much of the accuracy-driven literature.
We formulate landmark-based point-of-regard estimation as a session-wise adaptation problem in which a shared geometric encoder produces embeddings that can be aligned to a new session from only a small calibration set.
We present \emph{Equivariant Meta-Calibrated Gaze} (EMC-Gaze), a lightweight landmark-only method that combines an E(3)-equivariant landmark-graph encoder, richer local eye geometry, binocular eye emphasis, training-only auxiliary 3D gaze-direction supervision, and a closed-form ridge calibrator through which we differentiate during episodic meta-training.
To reduce pose leakage without introducing a high-capacity deployment-time pose pathway, we add a two-view canonicalization consistency loss and an optional low-capacity pose residual.
The deployed predictor consumes only derived facial landmarks and head pose and fits a tiny per-session ridge head from a brief calibration routine.
In a fixation-style interactive evaluation over 33 sessions at an approximate viewing distance of \SI{100}{cm}, EMC-Gaze achieves \SI{5.79 \pm 1.81}{\degree} overall angular RMSE after 9-point calibration, compared with \SI{6.68 \pm 2.34}{\degree} for Elastic Net, the strongest raw-landmark baseline.
The gain is larger on still-head queries (\SI{2.92 \pm 0.75}{\degree} versus \SI{4.45 \pm 0.30}{\degree}).
As our primary generalization benchmark, we report a retrospective subject-disjoint evaluation: across three randomized holdouts of 10 held-out subjects each, EMC-Gaze retains an advantage over Elastic Net (\SI{5.66 \pm 0.19}{\degree} versus \SI{6.49 \pm 0.33}{\degree} overall angular RMSE across split means).
On the public MPIIFaceGaze benchmark under leave-one-person-out evaluation with short per-session calibration, the eye-focused EMC-Gaze achieves \SI{8.82 \pm 1.21}{\degree} angular error at 16-shot calibration, ties Elastic Net at 1-shot, and outperforms it from 3-shot calibration onward.
The exported eye-focused encoder contains 944{,}423 parameters and occupies approximately \SI{4.76}{MB} as an ONNX artifact; full calibrated browser prediction requires 12.58/12.58/12.90 ms per sample (mean/median/p90) in headless Chromium~145 with threaded ONNX Runtime Web.
These results position EMC-Gaze as a browser-capable, calibration-friendly operating point for webcam gaze tracking rather than as a claim of universal state-of-the-art performance against heavier appearance-based systems.
\end{abstract}

\begin{keyword}
gaze tracking \sep point-of-regard estimation \sep landmark-based estimation \sep session adaptation \sep geometric deep learning \sep browser deployment \sep runtime efficiency
\end{keyword}

\end{frontmatter}

\ifreview
  \linenumbers
  \doublespacing
\else
  \setstretch{1.05}
\fi

\section{Introduction}
Webcam gaze tracking aims to infer a user's point of regard on a display from commodity cameras, enabling applications in accessibility, attentive interfaces, and human--computer interaction.
In practice, however, useful webcam systems are constrained not only by prediction error but also by calibration burden, robustness to head motion and session changes, runtime footprint, and browser feasibility.
Recent surveys emphasize that \emph{personal calibration} and \emph{head motion} remain two of the main unresolved issues in practical remote gaze estimation, especially when camera placement and viewing geometry vary across sessions \cite{hansen2010survey,liu2022survey,cheng2024review,lei2024endtoend,liu2024calibreview}.
These issues are particularly acute in everyday webcam use, where the same user may appear under different lighting, posture, and device placement conditions across runs.

Appearance-based gaze estimators have achieved strong results on public datasets and free-head settings such as MPIIGaze, GazeCapture, ETH-XGaze, and Gaze360, building on earlier work in synthesis, full-face modeling, data normalization, and monocular 3D tracking \cite{zhang2015mpiigaze,krafka2016eyetracking,zhang2020ethxgaze,kellnhofer2019gaze360,sugano2014synthesis,zhang2017fullface,zhang2018normalization,zhu2017monocular}.
Practical webcam systems and studies have additionally explored browser deployment, online experimentation, calibration efficiency, validation against lab-grade trackers, and screen-interaction settings \cite{papoutsaki2016webgazer,papoutsaki2017searchgazer,semmelmann2018firstlook,yang2021behavioral,gudi2020efficiency,saxena2024online,kaduk2024lab,valtakari2024field,vandercruyssen2024validation,falch2024screen,patterson2025scoping}.
Because browser/webcam studies differ substantially in head-motion control, calibration burden, viewing geometry, and runtime assumptions, cross-paper numerical comparisons should be interpreted cautiously \cite{semmelmann2018firstlook,saxena2024online,kaduk2024lab,valtakari2024field,patterson2025scoping}.
These lines of work suggest that raw benchmark accuracy is only one part of the deployment story.
Landmark-guided work has also shown that eye and facial geometry carry substantial gaze information \cite{park2018landmarks,yu2018multitask,sun2024semisup,gou2024cascaded}.
This makes a landmark-only pipeline attractive when a lightweight geometric representation is preferred over storing or processing raw facial imagery at inference.
However, landmark robustness is not automatic: rigid head motion directly changes the measured 3D configuration, so the representation must retain gaze-relevant geometry while suppressing nuisance Euclidean transforms.
Our goal is therefore not to outrun large raw-image backbones on unconstrained leaderboards, but to target a narrower regime in which a short explicit calibration routine is acceptable because it enables a smaller, more controllable, and browser-capable runtime.

Existing personalization methods confirm the value of rapid adaptation for gaze estimation \cite{park2019faze,linden2019personalize,he2019ondevice}.
We argue that calibration should be the \emph{primary adaptation operator} that shapes representation learning.
If deployment relies on a tiny per-session regressor fitted from a handful of still-head samples, then the shared encoder should be trained so that this exact low-capacity calibrator remains reliable under head motion and cross-session variability.

From this perspective, we introduce \emph{Equivariant Meta-Calibrated Gaze} (EMC-Gaze), a deployment-oriented landmark-only webcam gaze tracker built around four design choices.
First, we use an E(3)-equivariant landmark-graph encoder because the input is geometric and rigid head transforms are the dominant nuisance variation.
Second, we add richer local eye geometry, binocular eye emphasis, and a compact set of rotation-invariant iris-distance features to preserve fine ocular cues that graph pooling can dilute.
Third, we optimize the shared representation \emph{through} the closed-form ridge solution during episodic meta-training so that learning is aligned with the exact adaptation mechanism used at inference.
Fourth, we regularize the encoder with a two-view canonicalization consistency loss together with training-only auxiliary 3D gaze-direction supervision, so that pose robustness is encouraged in the shared embedding rather than delegated to a heavy deployment-time branch.

Beyond the core method, we describe a session-based data-collection pipeline with explicit head-motion phases and an optional smooth-pursuit block.
The full training and evaluation pipeline is implemented within the EyeTrax gaze-tracking library, which provides the common collection, calibration, benchmarking, and live-demo infrastructure used throughout this study.
The pursuit extension is not intended to replace fixation-based evaluation; rather, it complements it by exposing continuity and interpolation behavior that pointwise fixation metrics can obscure.

\paragraph{Contributions.}
\begin{itemize}[leftmargin=*]
  \item We study a \textbf{deployment-oriented operating point} for webcam gaze tracking: \textbf{landmark-only runtime inputs}, \textbf{short per-session calibration}, and a \textbf{browser-capable implementation path} within the EyeTrax system.
  \item We formulate landmark-based webcam gaze tracking as \textbf{session-wise meta-calibration}, where a \textbf{closed-form ridge calibrator} is part of the training graph and is used unchanged at deployment.
  \item We combine an \textbf{E(3)-equivariant landmark-graph encoder} with \textbf{richer local eye geometry}, \textbf{binocular eye emphasis}, \textbf{iris-distance invariants}, and a \textbf{two-view canonicalization consistency loss} to improve robustness to rigid head motion in a lightweight geometric representation.
  \item We provide both \textbf{accuracy} and \textbf{deployment-facing} evidence through a \textbf{retrospective subject-disjoint benchmark}, a \textbf{session-level interactive benchmark}, a \textbf{few-shot MPIIFaceGaze calibration curve}, and a \textbf{runtime characterization} covering model size, calibration burden, latency, and browser export.
\end{itemize}

\section{Related work}
\paragraph{Remote gaze estimation and webcam systems.}
Recent reviews divide gaze estimation into 2D mapping-based, 3D model-based, and appearance-based families, and highlight calibration and head motion as continuing challenges in remote settings \cite{hansen2010survey,liu2022survey,cheng2024review,lei2024endtoend,liu2024calibreview}.
Classical and appearance-based work spans learning-by-synthesis, full-face modeling, monocular free-head 3D estimation, data normalization, and large in-the-wild datasets such as MPIIGaze, GazeCapture, ETH-XGaze, RT-GENE, and Gaze360 \cite{sugano2014synthesis,zhang2015mpiigaze,krafka2016eyetracking,zhu2017monocular,zhang2017fullface,zhang2018normalization,fischer2018rtgene,zhang2020ethxgaze,kellnhofer2019gaze360,zhang2019evaluation}.
Methodological work on webcam-based eye tracking has progressed from early browser systems and feasibility studies to behavioral experiments, calibration studies, direct validations, and recent scoping recommendations \cite{papoutsaki2016webgazer,papoutsaki2017searchgazer,semmelmann2018firstlook,yang2021behavioral,gudi2020efficiency,saxena2024online,kaduk2024lab,valtakari2024field,vandercruyssen2024validation,falch2024screen,patterson2025scoping}.
Our work addresses a narrower operating point: short-session calibrated webcam tracking from derived 3D landmarks rather than raw image inputs.

\paragraph{Landmark-based and landmark-guided gaze estimation.}
Eye landmarks provide structured geometric cues for gaze inference.
Park et al.\ showed that learned eye-region landmarks can support competitive personalized and person-independent gaze estimation in unconstrained settings \cite{park2018landmarks}.
Subsequent work has used landmarks as explicit geometric constraints, auxiliary supervision, or collaborative multi-task signals to improve gaze estimation accuracy \cite{yu2018multitask,sun2024semisup,gou2024cascaded}.
Our method differs in that the runtime representation is itself a 3D facial-landmark graph, and the deployment-time predictor is a per-session screen-space calibrator fitted separately for each session.

\paragraph{Calibration and personalization.}
Calibration remains central in remote gaze tracking because person-specific anatomy and setup geometry induce systematic bias \cite{liu2024calibreview}.
Personalization and low-shot adaptation have been explored through latent parameterization, fine-tuning, and meta-learning, including FAZE, appearance-based personalization, and on-device few-shot personalization \cite{park2019faze,linden2019personalize,he2019ondevice}.
Model-based work has also sought to reduce or avoid explicit calibration under stronger geometric assumptions \cite{wang2018noexplicit}.
Our work shares the rapid-adaptation motivation but deliberately constrains adaptation to a closed-form ridge regressor that is identical during meta-training and deployment.

\paragraph{Meta-learning with closed-form adaptation.}
Differentiating through convex or closed-form adaptation steps is standard in few-shot learning, including methods with ridge-regression heads \cite{bertinetto2018closedform}.
We adopt this perspective for session-wise gaze calibration: each session is treated as a task, and the support-set ridge solution is part of the computation graph.

\paragraph{Equivariant representations for geometric inputs.}
Equivariant graph neural networks provide a principled way to process point sets under rigid transforms \cite{satorras2021egnn}.
In our setting, this inductive bias is appealing because the main nuisance factors are Euclidean motions of a 3D landmark configuration.
We combine an EGNN-style encoder with an auxiliary consistency objective that further discourages the embedding from entangling global head pose with gaze.

\paragraph{Smooth-pursuit paradigms.}
Smooth-pursuit interaction uses continuous eye motion induced by moving targets to support gaze-based interaction beyond static point selection \cite{vidal2013pursuits}.
We draw on this literature only for data collection and evaluation: pursuit segments provide dense supervision that reveals interpolation quality and temporal continuity beyond what discrete fixation targets alone can show.

\section{Problem formulation}
We observe a stream of webcam frames and aim to predict a 2D on-screen target location.
Instead of raw pixels, we extract 3D facial landmarks and a head pose estimate.
Data are organized into \emph{sessions}. A session corresponds to a single user/setup run containing timestamped samples with known on-screen targets.
At deployment, a new session is adapted using a small calibration support set.

Formally, a session provides $N$ samples:
\[
\{(X_n, p_n, y_n)\}_{n=1}^N,
\]
where $X_n \in \R^{3V}$ are 3D coordinates for a fixed subset of $V$ landmarks (flattened),
$p_n \in \R^3$ is head pose (yaw, pitch, roll),
and $y_n \in \R^2$ is the 2D on-screen target (normalized or pixels).

We learn (i) an encoder $f_\theta$ that maps a landmark graph to an embedding $e \in \R^d$ shared across sessions, and
(ii) a per-session linear regressor $W$ (ridge) mapping embeddings to $y$.
The central modeling assumption is that, once nuisance session factors are suppressed in the embedding, the remaining session-specific mapping to screen coordinates is simple enough to be captured by a regularized linear calibrator fitted from only a few samples.

\section{EMC-Gaze}
\subsection{Overview}
\label{sec:overview}
\begin{figure}[t]
\centering
\includegraphics[width=\linewidth]{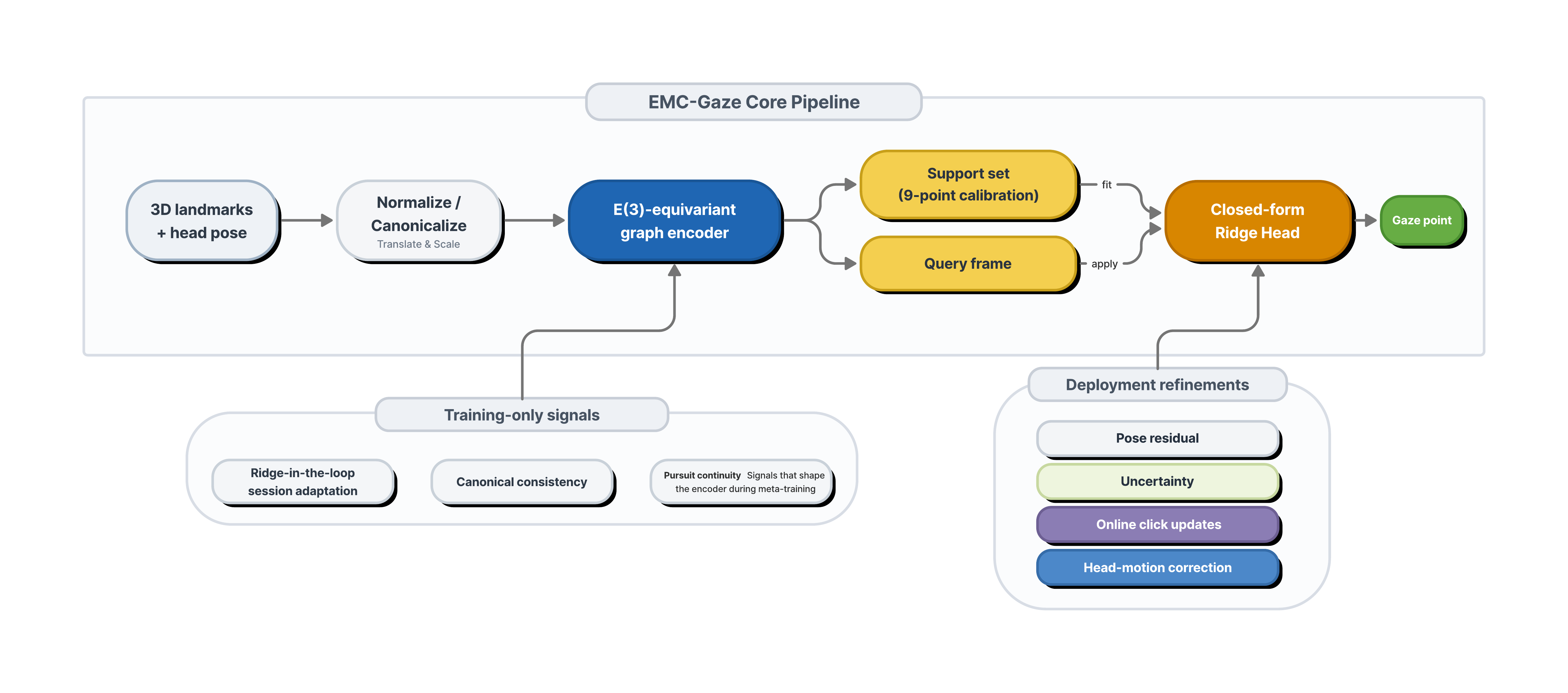}
\caption{\textbf{EMC-Gaze overview.} Landmark-only webcam gaze tracking is decomposed into a shared E(3)-equivariant landmark-graph encoder, a closed-form per-session ridge calibration head fitted on a short support set, and low-capacity deployment refinements. During meta-training, the encoder is optimized through the ridge solution together with canonicalization-consistency and pursuit-continuity regularization.}
\label{fig:method}
\end{figure}

EMC-Gaze decomposes gaze tracking into three parts: a shared equivariant encoder $f_\theta$, a per-session ridge regressor fitted on the calibration support set, and an optional low-capacity pose residual $g_\phi$.
This decomposition is deliberate.
Cross-session structure should be absorbed by the encoder, session-specific affine alignment should be handled by the calibrator, and explicit pose should be permitted only through a tightly constrained additive pathway.
In the flagship implementation, the encoder is explicitly eye-focused: capacity is concentrated on richer left/right eye geometry and binocular fusion, with the rest of the face retained only as lightweight stabilizing context.
During training only, we further attach a small auxiliary 3D gaze-direction head to the shared embedding.
This branch is supervised against a canonicalized direction proxy derived from the 2D targets, but it is discarded at deployment time so that the runtime predictor remains the same closed-form calibrated 2D head.
Throughout the remainder of this paper, ``EMC-Gaze'' refers to this eye-focused flagship implementation unless noted otherwise.
Figure~\ref{fig:method} summarizes the full EMC-Gaze pipeline, including the deployment-time calibration path and the training-only regularizers used to shape the shared representation.

\subsection{Landmark graph construction}
Nodes correspond to $V{=}161$ selected landmark indices from the MediaPipe face mesh: left eye region (76), right eye region (76), and a small set of mutual anchors (9).
Edges are derived from the FaceMesh tessellation adjacency and filtered to keep only edges whose endpoints are in the subset; the resulting graph is made bidirectional.
This subset concentrates capacity on the eye regions, where gaze cues are strongest, while retaining enough shared anchors to stabilize normalization and canonicalization across the full face geometry.

\subsection{E(3)-equivariant landmark-graph encoder}
We adopt an EGNN-style message passing architecture \cite{satorras2021egnn} that maintains
node coordinates $x_i \in \R^3$ (equivariant to Euclidean transforms) and node embeddings $h_i \in \R^d$ (intended to be invariant).
An equivariant encoder is a natural fit for this problem: the dominant nuisance transformations are rigid motions of a 3D landmark set, yet the downstream calibrator still benefits from local geometric structure that a purely handcrafted invariant representation might discard.
Our latest refinement further biases this encoder toward the eye regions: it augments the landmark graph with richer local eye-geometry features, pools the left and right eye regions explicitly, and fuses those binocular summaries with only a small face-context pathway.

\paragraph{Normalization.}
Regardless of how landmarks are stored, the model performs translation and scale normalization inside the forward pass:
translate to eye center and divide by inter-ocular distance.
This removes global translation and scale variation across sessions.
We deliberately do \emph{not} fully canonicalize the main prediction path here; instead, canonicalization is used in an auxiliary consistency branch (Section~\ref{sec:canon-cons}) so that pose leakage is penalized without forcing every prediction through a single estimated face basis.

\paragraph{Initialization.}
Coordinates are initialized from normalized landmarks; node features are initialized from a learned node identity embedding:
$h_i = \mathrm{Embed}(i)$.

\paragraph{Layer update.}
For each directed edge $(i \rightarrow j)$, define $d_{ij}=x_j-x_i$ and squared distance $r_{ij}^2=\|d_{ij}\|^2$.
Compute edge messages
\[
m_{ij} = \phi_e([h_i,h_j,r_{ij}^2]),
\]
and a coordinate weight
\[
w_{ij} = \tanh(\phi_x(m_{ij})).
\]
Aggregate coordinate updates:
\[
\Delta x_j = \frac{1}{\deg(j)+\epsilon} \sum_{i \in \mathcal{N}(j)} w_{ij} \, d_{ij},
\quad x_j \leftarrow x_j + \Delta x_j,
\]
and feature updates:
\[
m_j = \sum_{i \in \mathcal{N}(j)} m_{ij},
\quad h_j \leftarrow h_j + \phi_h([h_j,m_j]),
\]
with dropout applied in $\phi_h$.
A pooled node embedding (mean pooling) is combined with additional invariants (Section~\ref{sec:iris-inv}) and mapped to the final embedding $e$ via an MLP head.

\subsection{Invariant iris-distance features}
\label{sec:iris-inv}
Pure message passing can under-emphasize the fine relative displacement between iris centers and nearby eyelid or eye-corner anchors, even though these cues are strongly correlated with gaze.
We therefore compute a small set of rotation- and translation-invariant distances.
Left and right iris centers are estimated as the mean of iris landmarks, and distances from each iris center to fixed eyelid/corner anchors are concatenated with the pooled graph embedding before the final MLP head.
These low-dimensional invariants stabilize few-shot calibration rather than replace the learned representation.

\subsection{Per-session calibration head: ridge regression}
Given a support set $\{(e_k, y_k)\}_{k=1}^K$, we fit a ridge regressor with bias.
Let $H=[e\;\; \mathbf{1}] \in \R^{K \times (d+1)}$ and $Y \in \R^{K\times 2}$.
The closed-form solution is:
\begin{equation}
W = (H^\top H + \lambda I)^{-1} H^\top Y,
\end{equation}
where $W \in \R^{(d+1)\times 2}$.
For a query embedding $e$, prediction is $\hat{y} = [e\;\;1]W$.

The choice of ridge is deliberate.
With only a handful of calibration samples (typically a 9-point grid), a closed-form linear calibrator is statistically stable, computationally negligible, and exactly matches the adaptation mechanism used at inference.
The encoder is therefore trained to produce embeddings for which this low-capacity head is sufficient.

\subsection{Optional pose residual head}
If enabled, a small MLP $g_\phi:\R^3 \to \R^2$ maps $(\mathrm{yaw},\mathrm{pitch},\mathrm{roll})$ to a 2D correction.
During calibration we subtract this correction before fitting ridge:
$y_k^{\mathrm{eff}} = y_k - g_\phi(p_k)$,
fit ridge on $(e_k \rightarrow y_k^{\mathrm{eff}})$,
and at inference add it back:
\[
\hat{y} = [e\;\;1]W + g_\phi(p).
\]
Its role is intentionally limited: the pose residual can explain only an additive screen-space bias and cannot replace the landmark-to-gaze mapping learned by the main encoder.
This preserves the geometric inductive bias of the core model while still allowing a small correction for residual pose-dependent effects.

\subsection{Practical deployment extensions}
We additionally include several deployment-oriented engineering extensions.
These keep the encoder frozen and are reported for completeness rather than as primary methodological contributions.

\paragraph{Implicit online calibration.}
At deployment, the ridge head can be updated using occasional user click events as additional supervision while keeping the encoder fixed.
Each click contributes a new supervised pair $(e, y)$, and the head is updated with recursive least squares (online weighted ridge) using a forgetting factor.
To limit drift from noisy clicks, we retain the head estimated from the explicit calibration routine as a stable fallback and gradually blend toward the online-updated head only after sufficient evidence accumulates.

\paragraph{Inference-time head-motion correction.}
We also support an optional post-hoc correction layer based on relative 6-DoF head motion.
For each calibration frame, we estimate a compact head state
\[
h = (t_x, t_y, t_z, \mathrm{yaw}, \mathrm{pitch}, \mathrm{roll}),
\]
where $(t_x, t_y)$ are image-plane offsets of the eye center, $t_z$ is a scale-based depth proxy derived from inter-ocular distance, and the angular terms are computed from the face-attached landmark basis.
After fitting the base calibration head, we compute the residual error on the calibration frames and fit a small ridge regressor from relative head-state deltas $\Delta h$ to a 2D correction.
At inference, the final prediction becomes
\[
\hat{y}_{\mathrm{final}} = \hat{y}_{\mathrm{base}} + c(\Delta h),
\]
where $c(\cdot)$ is a bounded low-capacity correction estimated from the calibration set.
Because this correction is learned only from per-session residuals and clipped conservatively, it acts as a geometric refinement layer rather than a replacement for the gaze predictor.

\section{Meta-training}
\subsection{Episodic sessions and support/query splits}
We treat each session as a meta-learning task.
Each training episode samples one session and splits it into:
(i) a \textbf{support} (calibration) set consisting of still-head samples from the 9-point grid, and
(ii) a \textbf{query} set consisting of the remaining samples, including broader screen coverage and head-motion phases.
This construction deliberately mirrors deployment: calibrate on a short still-head routine and evaluate under more diverse conditions.
Restricting the support set to still-head samples is also a design choice: it prevents the calibrator from simply memorizing pose-specific corrections and forces robustness to head motion to emerge from the shared representation.

\subsection{Differentiable ridge-in-the-loop}
We compute the ridge solution within the training graph (e.g., using \texttt{torch.linalg.solve}),
so gradients backpropagate through the closed-form solver \cite{bertinetto2018closedform}.
This aligns training with inference: the encoder is optimized for the exact low-shot adaptation procedure that will be used at test time.

\subsection{Two-view canonicalization consistency}
\label{sec:canon-cons}
In addition to the query prediction loss, we apply an embedding consistency loss between a query sample and a canonicalized view of the same landmark coordinates.
Let $e_q=f_\theta(X_q)$ and $e_q^{\mathrm{canon}}=f_\theta(\mathcal{C}(X_q))$, where $\mathcal{C}$ rotates normalized landmarks into a canonical face basis.
The per-episode loss is:
\begin{equation}
\mathcal{L} =
\underbrace{\mathrm{MSE}(\hat{y}_Q, y_Q)}_{\text{query prediction}}
+
\lambda_{\mathrm{cons}}
\underbrace{\mathrm{MSE}(e_Q, e_Q^{\mathrm{canon}})}_{\text{two-view consistency}}.
\end{equation}
The main prediction path always uses the original normalized landmarks; canonicalization is applied only in this auxiliary branch.
This makes the regularizer tolerant to imperfect face-basis estimation while still discouraging the encoder from relying on global rigid head pose.

\subsection{Smooth-pursuit continuity loss}
Discrete fixation targets are sufficient to assess calibration accuracy, but they do not explicitly constrain whether predictions move smoothly between targets.
To address this, we introduce an optional continuity regularizer for pursuit segments.
When a query chunk contains temporally ordered smooth-pursuit samples with targets $\{y_t\}$ and predictions $\{\hat y_t\}$, we form frame-to-frame increments
\[
\Delta y_t = y_t - y_{t-1}, \qquad \Delta \hat y_t = \hat y_t - \hat y_{t-1},
\]
and add the loss
\[
\mathcal{L}_{\mathrm{cont}} = \mathrm{MSE}(\Delta \hat y_t, \Delta y_t).
\]
The full training objective becomes
\[
\mathcal{L}_{\mathrm{total}}
=
\mathcal{L}_{\mathrm{query}}
\;+\;
\lambda_{\mathrm{cons}} \mathcal{L}_{\mathrm{cons}}
\;+\;
\lambda_{\mathrm{cont}} \mathcal{L}_{\mathrm{cont}}.
\]
This term is active only on pursuit samples.
Its purpose is not to alter the fixation task, but to discourage the ``snapping'' behavior that can arise when the model is trained only on held targets and then evaluated on continuous motion.

\begin{algorithm}[t]
\caption{One EMC-Gaze meta-training episode}
\begin{algorithmic}[1]
\Require Session samples $\{(X_n,p_n,y_n)\}_{n=1}^N$; ridge $\lambda$; consistency weight $\lambda_{\mathrm{cons}}$; optional continuity weight $\lambda_{\mathrm{cont}}$.
\State Sample support indices $S$ (still head, grid stimuli) and query indices $Q$ (remaining).
\State $E_S \gets f_\theta(X_S)$; $E_Q \gets f_\theta(X_Q)$
\If{pose residual enabled}
  \State $G_S \gets g_\phi(p_S)$; $G_Q \gets g_\phi(p_Q)$
  \State $Y_S^{\mathrm{eff}} \gets Y_S - G_S$
\Else
  \State $G_Q \gets 0$; $Y_S^{\mathrm{eff}} \gets Y_S$
\EndIf
\State Fit ridge $W$ on $(E_S \rightarrow Y_S^{\mathrm{eff}})$ (closed form)
\State $\hat{Y}_Q \gets H_Q W + G_Q$
\State $E_Q^{\mathrm{canon}} \gets f_\theta(\mathcal{C}(X_Q))$
\State $\mathcal{L} \gets \mathrm{MSE}(\hat{Y}_Q,Y_Q) + \lambda_{\mathrm{cons}}\mathrm{MSE}(E_Q,E_Q^{\mathrm{canon}})$
\If{query contains pursuit samples}
  \State $\mathcal{L} \gets \mathcal{L} + \lambda_{\mathrm{cont}}\mathrm{MSE}(\Delta \hat{Y},\Delta Y)$
\EndIf
\State Update $(\theta,\phi)$ via backprop on $\mathcal{L}$
\end{algorithmic}
\end{algorithm}

\section{Data collection and feature extraction}
\subsection{Session-based protocol}
Each collection run is stored as a single \texttt{.npz} session file containing per-frame features, target coordinates, and metadata.
In the interactive evaluation used in this paper, each session contains 45 distinct stimulus targets:
a still-head 9-point calibration grid used as the \emph{support} set, plus 36 Poisson-disk targets used as \emph{queries}.
The query targets comprise 12 still-head Poisson points and 24 pose-hold points organized into six motion phases (\emph{yaw-left/right}, \emph{pitch-up/down}, \emph{roll-left/right}) with four Poisson targets per phase.
We store multiple samples per target (support: 5 samples/point; query: 2 samples/point), totaling 117 labeled frames per session.

\paragraph{Sessions, participants, and corpora.}
We use \emph{session} to denote one collected \texttt{.npz} run and \emph{participant} to denote one person, who may contribute multiple sessions.
The fixation-style training corpus contains 50 sessions from 32 participants.
Most participants contributed one session; two contributed three sessions each (seeds 0, 1, and 2), and one contributed 12 sessions in total: seven motion-enabled sessions (seeds 0--6) plus five still-only sessions collected with randomly sampled targets.
The main interactive benchmark contains 33 evaluation sessions from 33 evaluation participants.
These sessions were collected in a campus-recruited free-head setting with ordinary seating rather than a head-rest apparatus, which increases ecological variability while remaining a structured screen-interaction benchmark.
Because some evaluation participants also appear in the training corpus through separate sessions, the 33-run benchmark should be interpreted as a \emph{session-level} generalization test under new runs and conditions rather than as a fully participant-disjoint evaluation.
For the pursuit-enhanced EMC-Gaze variant, we additionally collected 10 smooth-pursuit sessions and trained encoder-based models on the combined 60-session corpus.
To obtain a stronger held-out-person assessment, we define the primary retrospective subject-disjoint protocol over the fixation-style corpus itself. Sessions from the author and two additional recurring participants are grouped directly; the March~6--7 follow-up sessions from the author are grouped with the same participant; and the March~5 train-then-evaluate collections are treated as distinct single-session subjects according to collection chronology.
This yields 35 reconstructed subject groups in total.
We then sample three randomized held-out sets of 10 March~5 subjects (10 sessions each) and train encoder-based models on the remaining 50 sessions for each split.

\paragraph{Optional smooth-pursuit extension.}
Motivated by smooth-pursuit paradigms in gaze interaction \cite{vidal2013pursuits}, the collection pipeline also supports an optional continuous target trajectory in which each saved frame is supervised by the instantaneous target position.
Our default pursuit path is a rounded serpentine scan that covers most of the screen while avoiding unnecessarily sharp turns; the implementation also supports figure-eight, horizontal, vertical, and Lissajous variants.
These pursuit samples are stored in the same session format with a dedicated phase label, so they can be used for continuity-aware training and for separate continuous-tracking evaluation without changing the underlying landmark representation.

\subsection{MediaPipe landmarks and input vector}
We use the MediaPipe framework and dense face-mesh pipeline to extract 3D face landmarks from video streams \cite{lugaresi2019mediapipe,kartynnik2019geometry}.
From the full face mesh, we retain $V{=}161$ selected landmarks (eyes + anchors) and concatenate head pose $(\mathrm{yaw},\mathrm{pitch},\mathrm{roll})$, yielding a per-frame feature vector of dimension $D = 3V + 3 = 486$ (pose in radians).

\subsection{Landmark canonicalization}
We support three landmark coordinate frames: \emph{raw}, \emph{centered} (translate to eye center and scale by inter-ocular distance), and \emph{canonical} (additionally rotate into a canonical face basis).
Canonicalization uses outer eye corners and a top-of-head anchor to build an orthonormal basis; details are provided in Appendix~\ref{app:canonical}.

\subsection{Blink filtering}
We compute a simple eye-aspect ratio (EAR) using eyelid landmarks with a short temporal history \cite{soukupova2016realtime}.
Frames below a dynamic threshold are marked as blinks and can be skipped.

\section{Experiments}
\subsection{Metrics}
We report RMSE for the 2D output and convert screen-plane error to an \emph{angular error} (degrees) using the viewing distance $d$:
$\theta = \arctan(\Delta/d)$,
where $\Delta$ is the screen-plane displacement in centimeters.
When targets are stored in normalized coordinates, we first convert to pixels and then to centimeters using the screen geometry.

\subsection{Evaluation protocol}
For each session we fit the ridge head using the still-head 9-point calibration grid
(support set: 9 targets $\times$ 5 samples/target $=45$ samples)
and evaluate on the remaining Poisson-disk query targets
(36 targets $\times$ 2 samples/target $=72$ samples).
The query set contains 12 still-head targets (24 samples) and 24 pose-hold targets (48 samples) spread across six motion phases (8 samples per phase).
We report angular RMSE for (i) still-head query samples, (ii) pose-hold blocks (mean across yaw/pitch/roll), and (iii) overall.
Because some evaluation participants also contributed separate training sessions, these metrics should be read as evidence for session-level robustness after short calibration, not as a definitive subject-independent benchmark.
For the retrospective subject-disjoint split, we use the same 9-point per-session support construction and report overall angular RMSE on the remaining query samples. Because the training-style held-out sessions do not include a separate still-head query block once the full support is consumed, we do not report a still/pose breakdown for that benchmark.

\paragraph{Benchmark summary.}
We therefore report two benchmarks with different goals. The \emph{primary benchmark} is a retrospective subject-disjoint protocol over the fixation-style corpus, which reports overall angular RMSE across three randomized held-out-subject splits. The \emph{complementary benchmark} is the 33-run session-level interactive evaluation, which reports still-head, pose-hold, and overall angular RMSE after the standard 9-point calibration. The subject-disjoint benchmark is intentionally narrower: because the held-out sessions come from the training-style collector and the full 9-point grid is consumed as support, no separate neutral-head query block remains for a still/pose decomposition.

\paragraph{Separate pursuit evaluation workflow.}
Because continuous-tracking quality is not fully captured by held-target accuracy, we additionally define a pursuit evaluation workflow.
Given sessions containing pursuit samples, this evaluator calibrates on the still 9-point support, predicts the pursuit trajectory, and reports both trajectory error and frame-to-frame motion error on the predicted path.
We do not yet include a large pursuit benchmark in the main fixation table, but this protocol is used both during model development and in the separate pursuit table reported below.

\subsection{Baselines}
We compare EMC-Gaze with:
(i) a ridge regressor on the raw landmark feature vector, included as a familiar calibrated linear baseline,
(ii) an Elastic Net regressor on the raw landmark feature vector, which serves as our stronger classical baseline, and
(iii) a learned graph encoder trained without the proposed equivariance or consistency components (``Meta GNN'') but using the same ridge calibration head.
All methods use the same landmark inputs, calibration routine, and evaluation protocol.

Because the paper studies the matched-input \emph{landmark-only} calibrated regime, we restrict direct numerical comparison to methods that operate on the same landmark and pose inputs under the same calibration and evaluation protocol.
This appropriately narrows the empirical claim: EMC-Gaze is evaluated as a geometric, session-adaptive alternative within the landmark-only setting rather than as a universal replacement for appearance-based gaze estimators.

\subsection{Results}
Table~\ref{tab:results} reports results under the session-level interactive protocol: 9-point still-head calibration followed by query targets under both neutral head pose and explicit yaw/pitch/roll hold blocks.
At evaluation time, we fit the per-session ridge head on the full 9-point support and measure angular RMSE on the remaining query samples.
These numbers correspond to the pursuit-enhanced EMC-Gaze benchmark suite collected before the final eye-focused refinement; the external MPIIFaceGaze benchmark and current deployment artifacts use the final eye-focused flagship described in Section~\ref{sec:overview}.
Among the matched landmark-input baselines, Elastic Net is the strongest conventional comparator, and EMC-Gaze improves on it for still-head, pose-hold, and overall error.
Because this benchmark has partial participant overlap between training and evaluation, we treat it as the session-level companion to the primary subject-disjoint analysis reported below.

\begin{table}[t]
\centering
\small
\setlength{\tabcolsep}{0pt}
\renewcommand{\arraystretch}{1.14}
\caption{Interactive evaluation across 33 runs at an approximate viewing distance of \SI{100}{cm}. Values report angular RMSE (degrees; mean$\pm$std) on the query split.}
\label{tab:results}
\begin{tabular*}{\linewidth}{@{\extracolsep{\fill}}L{0.46\linewidth}C{0.15\linewidth}C{0.15\linewidth}C{0.15\linewidth}@{}}
\toprule
Method & Still ($^\circ$) & Pose ($^\circ$) & Overall ($^\circ$) \\
\midrule
EMC-Gaze (ours, pursuit-enhanced) & \textbf{2.92$\pm$0.75} & \textbf{6.42$\pm$1.89} & \textbf{5.79$\pm$1.81} \\
Elastic Net & 4.45$\pm$0.30 & 7.11$\pm$2.49 & 6.68$\pm$2.34 \\
Ridge & 3.78$\pm$2.20 & 31.07$\pm$13.35 & 27.28$\pm$11.27 \\
Meta GNN & 3.68$\pm$2.20 & 29.72$\pm$11.98 & 26.41$\pm$10.18 \\
\bottomrule
\end{tabular*}
\end{table}

\paragraph{Primary subject-disjoint benchmark.}
To test whether the learned encoder remains useful on unseen people rather than only on unseen sessions, we additionally evaluate three randomized retrospective subject-disjoint splits constructed from collection chronology.
Each split holds out 10 reconstructed March~5 subjects (10 sessions) and retrains encoder-based models on the remaining 50 sessions while keeping the session-local calibration protocol unchanged.
Because the held-out sessions are drawn from the training-style corpus rather than the dedicated interactive evaluator, their query sets are dominated by motion conditions and do not support a separate still/pose breakdown after the full 9-point support has been consumed.
We emphasize this experiment because it provides the paper's strongest evidence that the learned representation transfers beyond reused participants.
Even under this stricter held-out-subject setting, EMC-Gaze remains better than the strongest classical baseline across all three splits.
EMC-Gaze outperformed Elastic Net on all three subject-disjoint splits, with a mean improvement of \SI{0.83}{\degree} angular RMSE, corresponding to an error reduction of roughly 13\%.
Meta GNN was included for completeness but was not competitive under this limited subject-disjoint training regime.
Additional checks with smaller and more strongly regularized Meta GNN variants did not improve held-out-subject performance, suggesting that the baseline transfers poorly in this regime rather than merely being poorly tuned.
For reproducibility, we save the frozen split manifests and split-specific checkpoints for this experiment, and the corresponding experiment code will be released as part of the EyeTrax codebase.

\begin{table}[t]
\centering
\small
\setlength{\tabcolsep}{0pt}
\renewcommand{\arraystretch}{1.14}
\caption{Primary retrospective subject-disjoint evaluation over 35 reconstructed subject groups. We sample three randomized held-out sets of 10 March~5 subjects (10 sessions) and retrain encoder-based models on the remaining 50 sessions for each split. Values report overall angular RMSE (degrees; mean$\pm$std across split means).}
\label{tab:subject_disjoint}
\begin{tabular*}{\linewidth}{@{\extracolsep{\fill}}L{0.62\linewidth}C{0.24\linewidth}@{}}
\toprule
Method & Held-out overall ($^\circ$) \\
\midrule
EMC-Gaze (ours, pursuit-enhanced) & \textbf{5.66$\pm$0.19} \\
Elastic Net & 6.49$\pm$0.33 \\
Ridge & 27.36$\pm$4.03 \\
Meta GNN & 33.23$\pm$3.33 \\
\bottomrule
\end{tabular*}
\end{table}

\paragraph{External public-dataset benchmark (MPIIFaceGaze LOPO).}
To test whether the learned representation transfers beyond the EyeTrax-collected corpus under a protocol that better matches EMC-Gaze's intended use, we evaluate on the public MPIIFaceGaze dataset with a leave-one-person-out (LOPO) protocol and short per-session calibration.
For each held-out participant, we train on the remaining 14 MPIIFaceGaze subjects, treat each subject-day as a session at test time, and fit the deployment-time ridge head from only $k \in \{1,3,5,9,16\}$ support samples chosen to cover the screen.
EyeTrax features are re-extracted from the MPIIFaceGaze images with the same MediaPipe landmark pipeline used elsewhere in this paper, and angular error is computed from the dataset's provided screen calibration and 3D face-center / gaze-target geometry.
Table~\ref{tab:mpiifacegaze} shows that the eye-focused EMC-Gaze matches Elastic Net at the extreme 1-shot setting and then pulls ahead from 3-shot calibration onward, with the gap widening as a practical few-shot calibration set becomes available.
At 16-shot calibration, the eye-focused EMC-Gaze reaches \SI{8.82 \pm 1.21}{\degree}, improving on Elastic Net by \SI{2.01}{\degree} (\SI{18.6}{\percent}).
We also tested a second setting that augments the MPIIFaceGaze training pool with 19 high-quality EyeTrax sessions collected on March~6--7 (still-only random coverage and pursuit-enhanced sessions); the resulting curve was effectively unchanged, so we report the cleaner MPII-only setting in the main table.

\begin{table}[t]
\centering
\footnotesize
\setlength{\tabcolsep}{3.5pt}
\renewcommand{\arraystretch}{1.12}
\caption{Leave-one-person-out MPIIFaceGaze benchmark with k-shot per-session calibration. Values report subject-macro angular RMSE (degrees; mean$\pm$std across the 15 held-out subjects).}
\label{tab:mpiifacegaze}
\resizebox{\linewidth}{!}{%
\begin{tabular}{lccccc}
\toprule
\multicolumn{1}{c}{Method} & \multicolumn{5}{c}{Calibration shots} \\
\cmidrule(lr){2-6}
 & 1 & 3 & 5 & 9 & 16 \\
\midrule
EMC-Gaze (ours, eye-focused, MPII-only) & \textbf{14.95$\pm$1.67} & \textbf{10.61$\pm$1.42} & \textbf{9.73$\pm$1.31} & \textbf{9.24$\pm$1.27} & \textbf{8.82$\pm$1.21} \\
Elastic Net & \textbf{14.95$\pm$1.67} & 11.72$\pm$1.41 & 11.15$\pm$1.31 & 10.95$\pm$1.31 & 10.83$\pm$1.31 \\
Ridge & \textbf{14.95$\pm$1.67} & 15.40$\pm$2.75 & 14.52$\pm$2.76 & 13.67$\pm$2.48 & 12.85$\pm$2.09 \\
\bottomrule
\end{tabular}%
}
\end{table}

\paragraph{Eye-focused refinement.}
After the main experiments above, we refined EMC-Gaze further by concentrating more of the encoder capacity on binocular eye geometry rather than broader face context.
On the internal 60-session EyeTrax benchmark, this eye-focused variant reduced normalized RMSE from 0.4699 to 0.4551.
Under the full leave-one-person-out MPIIFaceGaze protocol above, the same refinement improves the 16-shot result from 9.07$\pm$0.87 to 8.82$\pm$1.21 degrees, while also improving the 3-shot result from 11.64$\pm$1.32 to 10.61$\pm$1.42 degrees.
We therefore adopt the eye-focused encoder as the default EMC-Gaze implementation for the remainder of the paper.

\paragraph{Auxiliary 3D training refinement.}
We additionally found that 3D geometry is most useful when used as \emph{training-only supervision} rather than as a deployment-time bottleneck that the calibrated head must depend on directly.
Specifically, we add a small auxiliary 3D gaze-direction head on top of the shared embedding and train it against a canonicalized direction proxy, while leaving the deployment-time ridge calibration path unchanged.
On the internal 60-session EyeTrax benchmark, this refinement further improves normalized RMSE from 0.4551 to 0.4517, improves still-head RMSE from 0.1909 to 0.1852, and improves pursuit-phase RMSE from 0.2605 to 0.2572.
On a reduced MPIIFaceGaze sanity subset (8 subjects, 2 days each), the same refinement improves session-macro angular error from 9.83$\pm$1.03 to 9.57$\pm$1.02 degrees.
We therefore retain auxiliary 3D supervision in the flagship EMC-Gaze training recipe while keeping inference-time calibration purely 2D and closed-form.

\paragraph{Continuous pursuit tracking.}
The fixation benchmark above does not directly reveal whether predictions move smoothly and with the correct spatial amplitude between discrete targets.
Table~\ref{tab:pursuit} therefore reports the separate smooth-pursuit evaluation on 10 dedicated pursuit sessions.
EMC-Gaze yields the lowest trajectory RMSE after retraining on the same 60-session corpus, while remaining substantially better than Elastic Net and Meta GNN in overall trajectory accuracy.

\begin{table}[t]
\centering
\small
\setlength{\tabcolsep}{0pt}
\renewcommand{\arraystretch}{1.14}
\caption{Smooth-pursuit evaluation across 10 pursuit sessions at an approximate viewing distance of \SI{100}{cm}. Values report mean$\pm$std; lower is better except for the standard-deviation ratio, where values far below 1 indicate a collapsed trajectory.}
\label{tab:pursuit}
\begin{tabular*}{\linewidth}{@{\extracolsep{\fill}}L{0.37\linewidth}C{0.19\linewidth}C{0.18\linewidth}C{0.17\linewidth}@{}}
\toprule
Method & Trajectory RMSE ($^\circ$) & Delta RMSE ($^\circ$) & Std ratio (x/y) \\
\midrule
EMC-Gaze (ours, pursuit-enhanced) & \textbf{2.78$\pm$0.25} & 0.93$\pm$0.11 & 0.95 / 0.64 \\
Elastic Net & 4.86$\pm$0.08 & \textbf{0.57$\pm$0.02} & 0.22 / 0.20 \\
Ridge & 3.07$\pm$0.36 & 0.99$\pm$0.16 & 1.08 / 0.99 \\
Meta GNN & 3.15$\pm$0.53 & 1.08$\pm$0.15 & 1.04 / 1.03 \\
\bottomrule
\end{tabular*}
\end{table}

Elastic Net achieves the smallest frame-to-frame motion error, but Table~\ref{tab:pursuit} shows that it does so by severely shrinking the predicted trajectory range (standard-deviation ratios 0.22/0.20), which leads to substantially worse absolute tracking accuracy.
By contrast, EMC-Gaze preserves much more of the target trajectory amplitude while remaining accurate overall.
We therefore retain EMC-Gaze as the primary pursuit model because it offers the best trajectory RMSE together with a realistic dynamic range.

\paragraph{Ablations.}
Table~\ref{tab:ablations} reports a component ablation study for the core fixation-style EMC model.
Removing the canonicalization consistency loss mildly worsens overall accuracy.
Removing the pose residual head is essentially neutral on this corpus, suggesting that most robustness is already carried by the encoder and the session calibrator.
By contrast, removing the iris-distance invariant features or replacing the equivariant encoder with a simple MLP degrades performance, particularly under still-head conditions, indicating that both the geometric inductive bias and the added ocular invariants contribute meaningfully.

\begin{table}[t]
\centering
\small
\setlength{\tabcolsep}{0pt}
\renewcommand{\arraystretch}{1.14}
\caption{Ablations on the core fixation-style EMC model across 33 interactive runs. Values report angular RMSE (degrees; mean$\pm$std) on the query split.}
\label{tab:ablations}
\begin{tabular*}{\linewidth}{@{\extracolsep{\fill}}L{0.50\linewidth}C{0.12\linewidth}C{0.12\linewidth}C{0.12\linewidth}@{}}
\toprule
Variant & Still ($^\circ$) & Pose ($^\circ$) & Overall ($^\circ$) \\
\midrule
EMC-Gaze core (without pursuit extension) & 3.15$\pm$0.96 & 6.05$\pm$1.51 & 5.48$\pm$1.38 \\
No canonicalization consistency loss ($\lambda_{\mathrm{cons}}{=}0$) & 3.15$\pm$1.00 & 6.19$\pm$1.62 & 5.59$\pm$1.45 \\
No iris-distance invariant features & 5.19$\pm$0.04 & 5.93$\pm$0.05 & 5.72$\pm$0.04 \\
No pose residual head & 3.13$\pm$0.95 & 6.01$\pm$1.59 & 5.45$\pm$1.44 \\
Replace equivariant encoder with MLP encoder & 5.14$\pm$0.05 & 5.90$\pm$0.08 & 5.69$\pm$0.06 \\
\bottomrule
\end{tabular*}
\end{table}

\begin{figure}[t]
\centering
\includegraphics[width=\linewidth]{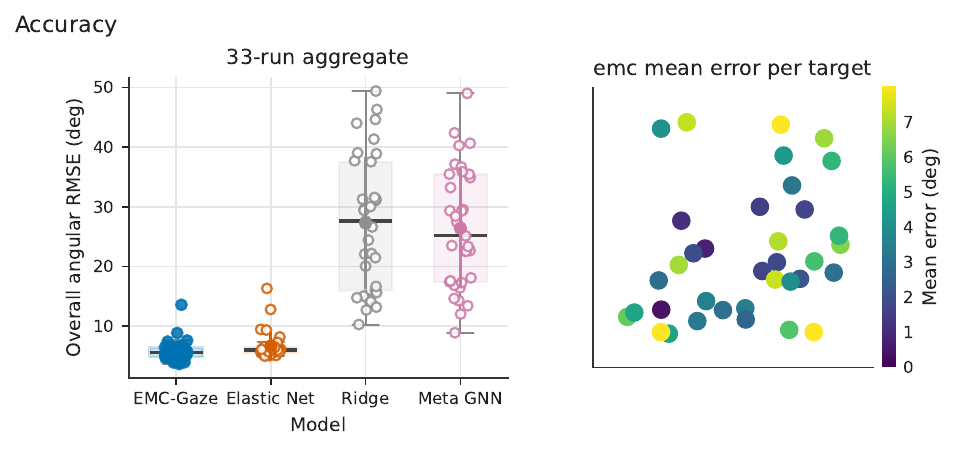}
\caption{\textbf{Accuracy.} (Left) Distribution of overall angular RMSE across the 33 interactive evaluation runs. Each point is one run; boxes summarize the run distribution. (Right) Mean angular error per target location (degrees) for EMC-Gaze on the representative run.}
\label{fig:accuracy}
\end{figure}

\begin{figure}[t]
\centering
\includegraphics[width=\linewidth]{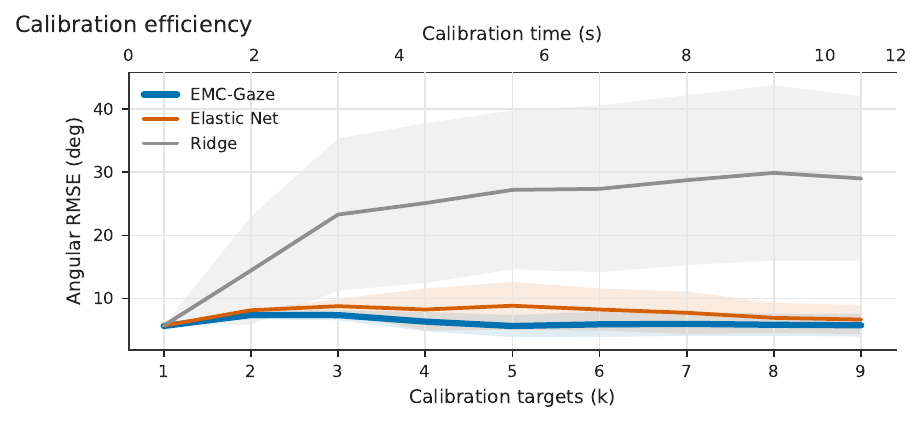}
\caption{\textbf{Efficient calibration.} Aggregate angular RMSE (degrees) versus number of calibration targets $k$ across the 33 interactive evaluation runs. Curves show mean performance; shaded bands show $\pm1$ standard deviation. The top axis reports approximate average calibration time for the EMC-Gaze prefix curve.}
\label{fig:calibration}
\end{figure}

\begin{figure}[t]
\centering
\includegraphics[width=\linewidth]{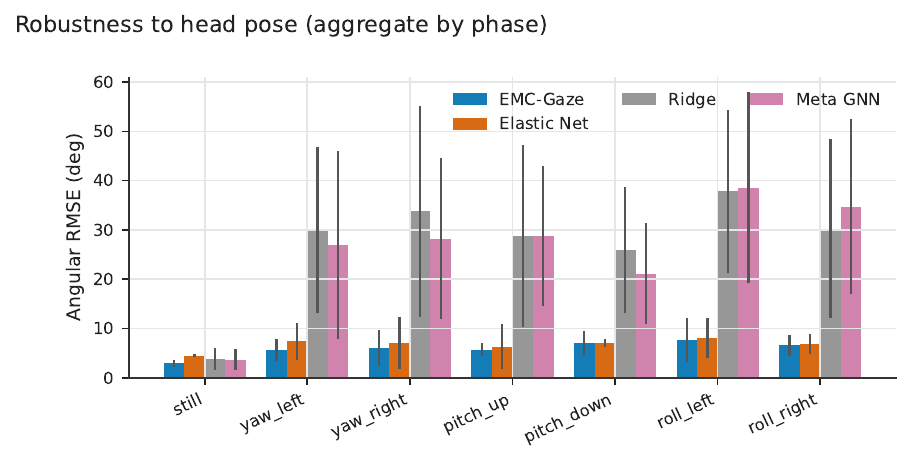}
\caption{\textbf{Robustness to head pose.} Mean per-phase angular RMSE (degrees) across the 33 interactive evaluation runs under yaw/pitch/roll pose-hold blocks (still-head calibration support only). Error bars denote $\pm1$ standard deviation across runs.}
\label{fig:robustness}
\end{figure}

\begin{figure}[t]
\centering
\includegraphics[width=\linewidth]{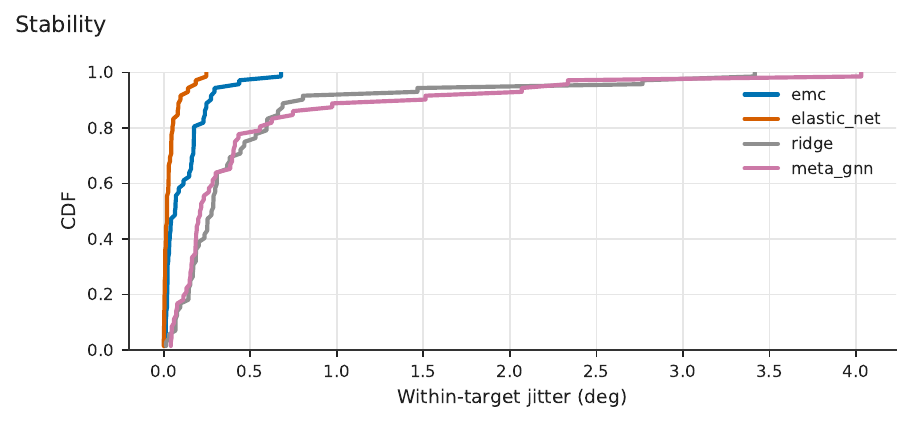}
\caption{\textbf{Stability.} CDF of within-target prediction jitter (degrees) on the representative run, computed as the distance from each prediction to that target's mean prediction. Lower is more stable.}
\label{fig:stability}
\end{figure}

\paragraph{Qualitative summaries.}
Figure~\ref{fig:accuracy}--Figure~\ref{fig:stability} summarize (i) aggregate accuracy across runs together with representative spatial error structure, (ii) calibration efficiency, (iii) robustness to head pose, and (iv) temporal stability for a representative interactive run.
Table~\ref{tab:results} provides the corresponding aggregate statistics across runs.

\section{Implementation details and deployment characteristics}
Because the paper targets a deployment-oriented operating point, we report not only the training configuration but also model size, calibration burden, and runtime latency.
A practical default configuration uses embedding dimension $d{=}32$, hidden dimension 128, 4 EGNN layers, dropout 0.10, learning rate $3\times 10^{-4}$, ridge $\lambda{=}10^{-3}$, a pose residual with hidden dimension 16, consistency weight $\lambda_{\mathrm{cons}}{=}0.10$, and a small auxiliary 3D gaze-direction head used only during training.
Episodes sample still-head support from the 9-point grid (typically 1--5 samples per calibration target) and subsample up to 256 query samples, optionally chunked for VRAM safety.

\paragraph{System and runtime.}
The reported experiments were run on an NVIDIA RTX 2000 Ada laptop GPU with MediaPipe~0.10.21.
All experiments were executed within the EyeTrax gaze-tracking library, which serves as the common implementation layer for session collection, model fitting, interactive evaluation, latency benchmarking, and live deployment.
The same flagship eye-focused encoder is also exported to a browser runtime that combines MediaPipe Tasks landmark extraction, ONNX Runtime Web inference, a user-facing calibration routine (explicit anchors plus straight-line pursuit), and optional click-based online calibration inside a static web page.
The exported eye-focused encoder contains 944{,}423 parameters and occupies approximately \SI{4.76}{MB} as an ONNX artifact; including the tiny 98-parameter pose-residual MLP yields a deployed calibrated core of 944{,}521 learned parameters, with the final session-specific ridge head fit analytically at runtime rather than stored as network weights.
A standard 10-epoch training run required approximately 30 minutes.
Under the main interactive evaluation protocol, each session uses 45 support samples from the still-head 9-point calibration grid and 72 query samples: 24 still-head query samples and 48 pose-hold query samples spread across six motion phases (8 samples per phase).
On a calibrated EMC-Gaze model measured over 120 valid frames, landmark extraction latency was 7.36/6.95/8.81 ms (mean/median/p90). Model inference latency was 1.94/1.92/2.02 ms, and end-to-end pipeline latency was 9.30/8.89/10.73 ms per frame.
To isolate deployment-time prediction latency apart from webcam capture and landmark extraction, we additionally replayed recorded feature vectors from a saved session using a 45-sample calibration fit and 120 query samples. On CPU, calibrated EMC-Gaze prediction required 8.96/9.33/10.18 ms per sample (mean/median/p90) in the Python runtime. In the exported browser runtime measured in headless Chromium~145 with cross-origin-isolated threaded ONNX Runtime Web (4 threads + proxy worker), encoder inference required 12.78/12.71/13.45 ms and full calibrated prediction required 12.58/12.58/12.90 ms per sample.
Table~\ref{tab:deployment-summary} consolidates the main deployment-facing characteristics of the flagship runtime configuration.

\begin{table}[t]
\centering
\caption{\textbf{Deployment summary for the flagship eye-focused EMC-Gaze runtime.} Values correspond to the exported browser-capable configuration discussed in the text.}
\label{tab:deployment-summary}
\begin{tabular}{p{0.38\linewidth}p{0.56\linewidth}}
\toprule
Characteristic & Value \\
\midrule
Exported encoder & 944{,}423 parameters; \SI{4.76}{MB} ONNX artifact \\
Deployed calibrated core & 944{,}521 learned parameters including the 98-parameter pose residual \\
Runtime calibration head & Closed-form ridge regressor fit per session at runtime \\
Calibration burden & 9 targets, 45 still-head support samples in the main interactive protocol \\
Browser demo calibration & User-facing anchor-plus-pursuit route with optional click-based online updates \\
Runtime stack & MediaPipe Tasks, ONNX Runtime Web, client-side ridge calibration/inference \\
Live GPU pipeline (mean/median/p90) & Landmarks 7.36/6.95/8.81 ms; inference 1.94/1.92/2.02 ms; end-to-end 9.30/8.89/10.73 ms \\
CPU calibrated prediction & 8.96/9.33/10.18 ms per sample in Python (feature replay) \\
Browser calibrated prediction & 12.58/12.58/12.90 ms per sample in headless Chromium~145 with threaded ONNX Runtime Web \\
\bottomrule
\end{tabular}
\end{table}

\section{Discussion, limitations, and ethics}
\paragraph{Deployment focus.}
EMC-Gaze should be read as a deployment-oriented operating point rather than as a universal leaderboard challenge.
The runtime input is a face-landmark graph rather than raw eye or face crops, adaptation is a tiny per-session ridge head, and the exported eye-focused encoder is small enough to support browser-side inference with client-side calibration.
In this regime, the relevant tradeoff is among post-calibration error, calibration burden, robustness to head motion, runtime footprint, and implementation simplicity.

\paragraph{Why the method is robust to head motion.}
Three design choices work together.
First, translation and scale normalization remove straightforward nuisance variation before message passing.
Second, the equivariant encoder and two-view canonicalization consistency loss discourage rigid-motion leakage into the embedding.
Third, because calibration support is collected under still-head conditions, the ridge head cannot simply memorize pose-conditioned corrections; the shared representation must remain usable on the pose-hold query blocks.
When enabled, the pose residual head provides only a small additive correction rather than a high-capacity shortcut.

\paragraph{Limitations.}
EMC-Gaze depends on reliable landmark extraction and may degrade under occlusion, extreme lighting, motion blur, or landmark detector failure.
The method predicts \emph{2D screen location} rather than a full 3D gaze vector.
Per-session calibration is required for best performance, and the meta-training objective assumes that calibration and deployment distributions are reasonably matched by the support/query split.
Our current interactive evaluation set covers 33 runs, and the pursuit benchmark covers 10 dedicated smooth-pursuit sessions; these numbers are sufficient to reveal meaningful qualitative and quantitative differences, but remain limited in scale and demographic diversity.
The 33-run interactive benchmark still contains partial participant overlap between training and evaluation, so it should be interpreted as evidence of session-level robustness after short calibration.
The paper's stronger transfer evidence comes from the three randomized retrospective subject-disjoint splits, although a prospectively annotated participant-disjoint benchmark would be stronger still.
The deployment-time online calibration and head-motion correction extensions are useful engineering additions to the system, but they are not isolated in the formal evaluation reported here.
Relative to head-rest or chin-rest webcam assays, the internal interactive benchmark is lower-control and free-head, but it remains a structured screen-interaction protocol rather than a fully in-the-wild study.
Finally, the pursuit extension has not yet been scaled to the same breadth as the fixation-style protocol.

\paragraph{Positioning relative to broader webcam gaze literature.}
Recent appearance-based and webcam-oriented systems trained on large public datasets report strong performance when raw images and large-scale supervision are available, and recent methodological papers continue to improve the practical quality of browser-based tracking \cite{cheng2024review,lei2024endtoend,saxena2024online,kaduk2024lab,valtakari2024field,vandercruyssen2024validation,falch2024screen,patterson2025scoping,xia2025ccl}.
Direct numerical comparison across browser/webcam papers should nevertheless be interpreted cautiously, because reported accuracy depends strongly on head-motion control, calibration burden, viewing geometry, and whether inference is browser-native, offline, or otherwise deployment-constrained \cite{saxena2024online,kaduk2024lab,valtakari2024field,patterson2025scoping}.
EMC-Gaze does not attempt to outrun these systems on their own terms.
Instead, it targets a narrower regime in which a short explicit calibration routine is acceptable and the runtime is intentionally lightweight, geometry-based, and exportable to the web.
The contribution should therefore be interpreted as methodological progress within the landmark-only calibrated regime and as evidence for an accuracy--deployability tradeoff, not as a claim of universal state-of-the-art performance.

\paragraph{Privacy and consent.}
Although the pipeline can store only derived facial landmarks and head pose (rather than raw video), these signals still constitute biometric data and should be collected with informed consent and appropriate data governance.
When collecting new datasets, we recommend documenting consent procedures, retention policies, and, when applicable, demographic balance and fairness analyses.

\section{Conclusion}
We presented EMC-Gaze as a deployment-oriented landmark-only webcam gaze tracker for the regime where a brief explicit calibration routine is acceptable but runtime simplicity, browser feasibility, and robustness to session changes matter.
The central methodological idea is to learn a shared geometric representation for the same closed-form ridge operator used at deployment, rather than optimizing an offline predictor and treating calibration as a separate afterthought.
Combined with an E(3)-equivariant landmark-graph encoder, richer local eye geometry, binocular eye emphasis, iris-distance invariants, canonicalization consistency, and training-only auxiliary 3D supervision, this yields a lightweight calibrated system that improves on matched landmark baselines after short calibration and transfers to held-out subjects and MPIIFaceGaze few-shot evaluation.
Because the deployed predictor consumes only derived landmarks together with a tiny session-specific ridge head, the same eye-focused encoder can be exported to a browser runtime with client-side calibration and inference.
The main contribution is therefore not a universal state-of-the-art claim against heavy appearance-based models, but a practical point on the tradeoff between accuracy, calibration burden, runtime footprint, and deployability for webcam gaze tracking.

\section*{Data availability}
The dataset supporting the findings of this study is not publicly released because it contains human-participant facial landmark and head-pose recordings that are treated as potentially identifying biometric data.
The code and trained model used for the reported experiments will be released alongside the public manuscript preprint.

\section*{Funding}
This research received no specific grant from any funding agency in the public, commercial, or not-for-profit sectors.

\section*{CRediT authorship contribution statement}
Chenkai Zhang: Conceptualization, Methodology, Software, Investigation, Formal analysis, Visualization, Writing -- original draft, Writing -- review \& editing.

\section*{Declaration of competing interest}
The author declares that there are no competing financial interests or personal relationships that could have appeared to influence the work reported in this paper.

\section*{Ethics statement}
All participants provided informed consent before data collection.
The study was conducted as a small independent non-institutional study by the sole author.
No formal ethics review board was available to review the study.

\appendix

\section{Canonicalization details}
\label{app:canonical}
Let $p_{33}$ and $p_{263}$ denote the outer eye corners and $p_{10}$ a top-of-head anchor (MediaPipe indices).
Define eye center $c=(p_{33}+p_{263})/2$ and inter-ocular distance $s=\|p_{263}-p_{33}\|$.
Centered points are $p_i \leftarrow (p_i - c)/(s+\epsilon)$.

For canonicalization, define:
\begin{align}
\hat{x} &= \mathrm{normalize}(p_{263}-p_{33}), \\
\hat{y}_0 &= \mathrm{normalize}(p_{10}-c), \\
\hat{y} &= \mathrm{normalize}(\hat{y}_0 - \langle \hat{y}_0, \hat{x}\rangle \hat{x}), \\
\hat{z} &= \mathrm{normalize}(\hat{x}\times \hat{y}),
\end{align}
and rotation matrix $R=[\hat{x}\; \hat{y}\; \hat{z}]$ (columns). Canonical coordinates are $p'_i = p_i R$.

\section{Supporting figures for a representative interactive run}
We include additional analyses used in the paper figures and ablations.

\begin{figure}[t]
\centering
\includegraphics[width=0.85\linewidth]{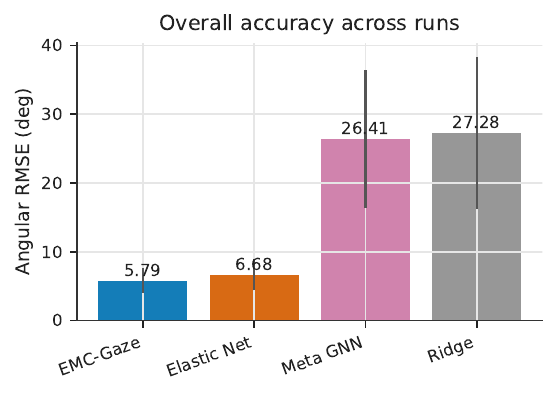}
\caption{\textbf{Overall accuracy comparison.} Mean overall angular RMSE (degrees) across the 33 interactive evaluation runs, with error bars indicating $\pm1$ standard deviation.}
\label{fig:supp-acc-bar}
\end{figure}

\begin{figure}[t]
\centering
\includegraphics[width=\linewidth]{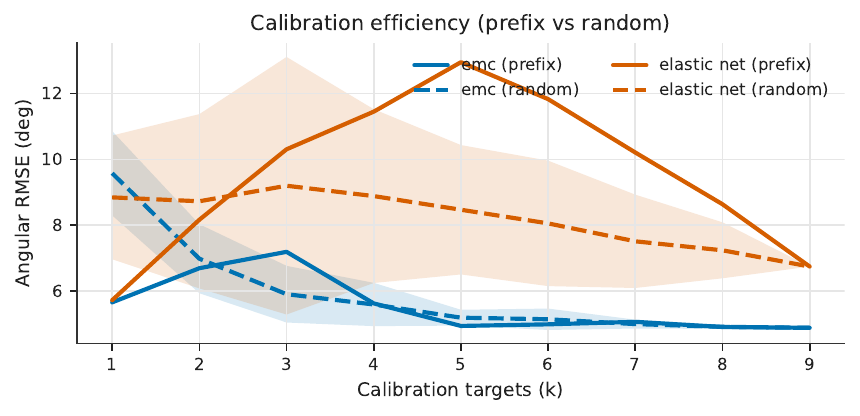}
\caption{\textbf{Calibration efficiency (prefix vs random).} Representative-run comparison in which prefix order uses the first $k$ calibration targets and random averages over random subsets (mean$\pm$std).}
\label{fig:supp-calib-full}
\end{figure}

\bibliographystyle{elsarticle-num}
\bibliography{references}

@inproceedings{krafka2016eyetracking,
  author    = {Kyle Krafka and Aditya Khosla and Petr Kellnhofer and Harini Kannan and Suchendra Bhandarkar and Wojciech Matusik and Antonio Torralba},
  title     = {Eye Tracking for Everyone},
  booktitle = {Proceedings of the IEEE Conference on Computer Vision and Pattern Recognition (CVPR)},
  pages     = {2176--2184},
  year      = {2016},
  doi       = {10.1109/CVPR.2016.239}
}

@inproceedings{zhang2015mpiigaze,
  author    = {Xucong Zhang and Yusuke Sugano and Mario Fritz and Andreas Bulling},
  title     = {Appearance-Based Gaze Estimation in the Wild},
  booktitle = {Proceedings of the IEEE Conference on Computer Vision and Pattern Recognition (CVPR)},
  pages     = {4511--4520},
  year      = {2015},
  doi       = {10.1109/CVPR.2015.7299081}
}

@inproceedings{zhang2020ethxgaze,
  author    = {Xucong Zhang and Seonwook Park and Thabo Beeler and Derek Bradley and Siyu Tang and Otmar Hilliges},
  title     = {{ETH-XGaze}: A Large Scale Dataset for Gaze Estimation under Extreme Head Pose and Gaze Variation},
  booktitle = {Computer Vision -- ECCV 2020},
  pages     = {365--381},
  year      = {2020},
  doi       = {10.1007/978-3-030-58558-7_22}
}

@inproceedings{satorras2021egnn,
  author    = {Victor Garcia Satorras and Emiel Hoogeboom and Max Welling},
  title     = {{E(n) Equivariant Graph Neural Networks}},
  booktitle = {Proceedings of the 38th International Conference on Machine Learning (ICML)},
  pages     = {9323--9332},
  year      = {2021}
}

@misc{bertinetto2018closedform,
  author    = {Luca Bertinetto and Jo{\~a}o F. Henriques and Philip H. S. Torr and Andrea Vedaldi},
  title     = {Meta-Learning with Differentiable Closed-Form Solvers},
  howpublished = {International Conference on Learning Representations (ICLR)},
  year      = {2019},
  url       = {https://openreview.net/forum?id=HyxnZh0ct7}
}

@article{lugaresi2019mediapipe,
  author  = {Camillo Lugaresi and Jiuqiang Tang and Hadon Nash and Chris McClanahan and Esha Uboweja and Michael Hays and Fan Zhang and Chuo{-}Ling Chang and Ming Guang Yong and Juhyun Lee and Wan{-}Teh Chang and Wei Hua and Manfred Georg and Matthias Grundmann},
  title   = {{MediaPipe}: A Framework for Building Perception Pipelines},
  journal = {arXiv preprint arXiv:1906.08172},
  year    = {2019}
}

@article{kartynnik2019geometry,
  author  = {Yury Kartynnik and Artsiom Ablavatski and Ivan Grishchenko and Matthias Grundmann},
  title   = {Real-Time Facial Surface Geometry from Monocular Video on Mobile GPUs},
  journal = {arXiv preprint arXiv:1907.06724},
  year    = {2019}
}

@inproceedings{park2019faze,
  author    = {Seonwook Park and Shalini De Mello and Pavlo Molchanov and Umar Iqbal and Otmar Hilliges and Jan Kautz},
  title     = {Few-Shot Adaptive Gaze Estimation},
  booktitle = {Proceedings of the IEEE/CVF International Conference on Computer Vision (ICCV)},
  pages     = {9368--9377},
  year      = {2019},
  doi       = {10.1109/ICCV.2019.00946}
}

@inproceedings{vidal2013pursuits,
  author    = {M{\'e}lodie Vidal and Andreas Bulling and Hans Gellersen},
  title     = {Pursuits: Spontaneous Interaction with Displays Based on Smooth Pursuit Eye Movement and Moving Targets},
  booktitle = {Proceedings of the 2013 ACM International Joint Conference on Pervasive and Ubiquitous Computing (UbiComp)},
  pages     = {439--448},
  year      = {2013},
  doi       = {10.1145/2493432.2493477}
}

@inproceedings{soukupova2016realtime,
  author    = {Tereza Soukupov{\'a} and Jan {\v{C}}ech},
  title     = {Real-Time Eye Blink Detection Using Facial Landmarks},
  booktitle = {21st Computer Vision Winter Workshop},
  pages     = {1--8},
  year      = {2016}
}

@article{hansen2010survey,
  author    = {Dan Witzner Hansen and Qiang Ji},
  title     = {In the Eye of the Beholder: A Survey of Models for Eyes and Gaze},
  journal   = {IEEE Transactions on Pattern Analysis and Machine Intelligence},
  volume    = {32},
  number    = {3},
  pages     = {478--500},
  year      = {2010},
  doi       = {10.1109/TPAMI.2009.30}
}

@article{liu2022survey,
  author    = {Jiahui Liu and Jiannan Chi and Huijie Yang and Xucheng Yin},
  title     = {In the Eye of the Beholder: A Survey of Gaze Tracking Techniques},
  journal   = {Pattern Recognition},
  volume    = {132},
  pages     = {108944},
  year      = {2022},
  doi       = {10.1016/j.patcog.2022.108944}
}

@article{cheng2024review,
  author    = {Yihua Cheng and Haofei Wang and Yiwei Bao and Feng Lu},
  title     = {Appearance-Based Gaze Estimation With Deep Learning: A Review and Benchmark},
  journal   = {IEEE Transactions on Pattern Analysis and Machine Intelligence},
  volume    = {46},
  number    = {12},
  pages     = {7509--7528},
  year      = {2024},
  doi       = {10.1109/TPAMI.2024.3393571}
}

@article{liu2024calibreview,
  author    = {Jiahui Liu and Jiannan Chi and Zuoyun Yang},
  title     = {A Review on Personal Calibration Issues for Video-Oculographic-Based Gaze Tracking},
  journal   = {Frontiers in Psychology},
  volume    = {15},
  pages     = {1309047},
  year      = {2024},
  doi       = {10.3389/fpsyg.2024.1309047}
}

@inproceedings{papoutsaki2016webgazer,
  author    = {Alexandra Papoutsaki and Patsorn Sangkloy and James Laskey and Nediyana Daskalova and Jeff Huang and James Hays},
  title     = {WebGazer: Scalable Webcam Eye Tracking Using User Interactions},
  booktitle = {Proceedings of the Twenty-Fifth International Joint Conference on Artificial Intelligence (IJCAI)},
  pages     = {3839--3845},
  year      = {2016}
}

@inproceedings{papoutsaki2017searchgazer,
  author    = {Alexandra Papoutsaki and James Laskey and Jeff Huang},
  title     = {SearchGazer: Webcam Eye Tracking for Remote Studies of Web Search},
  booktitle = {Proceedings of the 2017 Conference on Human Information Interaction and Retrieval (CHIIR)},
  pages     = {17--26},
  year      = {2017},
  doi       = {10.1145/3020165.3020170}
}

@incollection{gudi2020efficiency,
  author    = {Amogh Gudi and Xin Li and Jan van Gemert},
  title     = {Efficiency in Real-Time Webcam Gaze Tracking},
  booktitle = {Computer Vision -- ECCV 2020 Workshops},
  pages     = {529--543},
  publisher = {Springer},
  year      = {2021},
  doi       = {10.1007/978-3-030-66415-2_34}
}

@inproceedings{park2018landmarks,
  author    = {Seonwook Park and Xucong Zhang and Andreas Bulling and Otmar Hilliges},
  title     = {Learning to Find Eye Region Landmarks for Remote Gaze Estimation in Unconstrained Settings},
  booktitle = {Proceedings of the 2018 ACM Symposium on Eye Tracking Research \& Applications (ETRA)},
  articleno = {21},
  pages     = {1--10},
  year      = {2018},
  doi       = {10.1145/3204493.3204545}
}

@inproceedings{he2019ondevice,
  author    = {Junfeng He and Khoi Pham and Nachiappan Valliappan and Pingmei Xu and Chase Riley Roberts and Dmitry Lagun and Vidhya Navalpakkam},
  title     = {On-Device Few-Shot Personalization for Real-Time Gaze Estimation},
  booktitle = {Proceedings of the IEEE/CVF International Conference on Computer Vision Workshops (ICCVW)},
  pages     = {1149--1158},
  year      = {2019},
  doi       = {10.1109/ICCVW.2019.00146}
}

@article{sun2024semisup,
  author    = {Yunjia Sun and Jiabei Zeng and Shiguang Shan},
  title     = {Gaze Estimation with Semi-Supervised Eye Landmark Detection as an Auxiliary Task},
  journal   = {Pattern Recognition},
  volume    = {146},
  pages     = {109980},
  year      = {2024},
  doi       = {10.1016/j.patcog.2023.109980}
}

@article{gou2024cascaded,
  author    = {Chao Gou and Yuezhao Yu and Zipeng Guo and Chen Xiong and Ming Cai},
  title     = {Cascaded Learning with Transformer for Simultaneous Eye Landmark, Eye State and Gaze Estimation},
  journal   = {Pattern Recognition},
  volume    = {156},
  pages     = {110760},
  year      = {2024},
  doi       = {10.1016/j.patcog.2024.110760}
}

@article{falch2024screen,
  author    = {Lucas Falch and Katrin Solveig Lohan},
  title     = {Webcam-Based Gaze Estimation for Computer Screen Interaction},
  journal   = {Frontiers in Robotics and AI},
  volume    = {11},
  pages     = {1369566},
  year      = {2024},
  doi       = {10.3389/frobt.2024.1369566}
}

@article{saxena2024online,
  author    = {Shreshth Saxena and Lauren K. Fink and Elke B. Lange},
  title     = {Deep Learning Models for Webcam Eye Tracking in Online Experiments},
  journal   = {Behavior Research Methods},
  volume    = {56},
  pages     = {3487--3503},
  year      = {2024},
  doi       = {10.3758/s13428-023-02190-6}
}

@article{wang2018noexplicit,
  author    = {Kang Wang and Qiang Ji},
  title     = {3D Gaze Estimation without Explicit Personal Calibration},
  journal   = {Pattern Recognition},
  volume    = {79},
  pages     = {216--227},
  year      = {2018},
  doi       = {10.1016/j.patcog.2018.01.031}
}

@article{xia2025ccl,
  author    = {Lifan Xia and Yong Li and Xin Cai and Zhen Cui and Chunyan Xu and Antoni B. Chan},
  title     = {Collaborative Contrastive Learning for Cross-Domain Gaze Estimation},
  journal   = {Pattern Recognition},
  volume    = {161},
  pages     = {111244},
  year      = {2025},
  doi       = {10.1016/j.patcog.2024.111244}
}

@inproceedings{sugano2014synthesis,
  author    = {Yusuke Sugano and Yasuyuki Matsushita and Yoichi Sato},
  title     = {Learning-by-Synthesis for Appearance-Based 3D Gaze Estimation},
  booktitle = {Proceedings of the IEEE Conference on Computer Vision and Pattern Recognition (CVPR)},
  pages     = {1821--1828},
  year      = {2014},
  doi       = {10.1109/CVPR.2014.235}
}

@inproceedings{zhu2017monocular,
  author    = {Haoping Deng and Wangjiang Zhu},
  title     = {Monocular Free-Head 3D Gaze Tracking with Deep Learning and Geometry Constraints},
  booktitle = {Proceedings of the IEEE International Conference on Computer Vision (ICCV)},
  pages     = {3162--3171},
  year      = {2017},
  doi       = {10.1109/ICCV.2017.341}
}

@inproceedings{zhang2017fullface,
  author    = {Xucong Zhang and Yusuke Sugano and Mario Fritz and Andreas Bulling},
  title     = {It's Written All Over Your Face: Full-Face Appearance-Based Gaze Estimation},
  booktitle = {Proceedings of the IEEE Conference on Computer Vision and Pattern Recognition Workshops (CVPRW)},
  pages     = {2299--2308},
  year      = {2017},
  doi       = {10.1109/CVPRW.2017.284}
}

@inproceedings{zhang2018normalization,
  author    = {Xucong Zhang and Yusuke Sugano and Andreas Bulling},
  title     = {Revisiting Data Normalization for Appearance-Based Gaze Estimation},
  booktitle = {Proceedings of the 2018 ACM Symposium on Eye Tracking Research \& Applications (ETRA)},
  pages     = {12:1--12:9},
  year      = {2018},
  doi       = {10.1145/3204493.3204548}
}

@inproceedings{fischer2018rtgene,
  author    = {Tobias Fischer and Hyung Jin Chang and Yiannis Demiris},
  title     = {RT-GENE: Real-Time Eye Gaze Estimation in Natural Environments},
  booktitle = {Proceedings of the European Conference on Computer Vision (ECCV)},
  pages     = {339--357},
  year      = {2018},
  doi       = {10.1007/978-3-030-01249-6_21}
}

@incollection{yu2018multitask,
  author    = {Yu Yu and Gang Liu and Jean-Marc Odobez},
  title     = {Deep Multitask Gaze Estimation with a Constrained Landmark-Gaze Model},
  booktitle = {Computer Vision -- ECCV 2018 Workshops},
  pages     = {456--474},
  year      = {2018},
  publisher = {Springer},
  doi       = {10.1007/978-3-030-11012-3_35}
}

@article{semmelmann2018firstlook,
  author    = {Kilian Semmelmann and Sarah Weigelt},
  title     = {Online Webcam-Based Eye Tracking in Cognitive Science: A First Look},
  journal   = {Behavior Research Methods},
  volume    = {50},
  number    = {2},
  pages     = {451--465},
  year      = {2018},
  doi       = {10.3758/s13428-017-0913-7}
}

@inproceedings{kellnhofer2019gaze360,
  author    = {Petr Kellnhofer and Adri{\`a} Recasens and Simon Stent and Wojciech Matusik and Antonio Torralba},
  title     = {Gaze360: Physically Unconstrained Gaze Estimation in the Wild},
  booktitle = {Proceedings of the IEEE/CVF International Conference on Computer Vision (ICCV)},
  pages     = {6911--6920},
  year      = {2019},
  doi       = {10.1109/ICCV.2019.00701}
}

@inproceedings{zhang2019evaluation,
  author    = {Xucong Zhang and Yusuke Sugano and Andreas Bulling},
  title     = {Evaluation of Appearance-Based Methods and Implications for Gaze-Based Applications},
  booktitle = {Proceedings of the 2019 CHI Conference on Human Factors in Computing Systems},
  pages     = {1--13},
  year      = {2019},
  doi       = {10.1145/3290605.3300646}
}

@inproceedings{linden2019personalize,
  author    = {Erik Lind{\'e}n and Jonas Sj{\"o}strand and Alexandre Prouti{\`e}re},
  title     = {Learning to Personalize in Appearance-Based Gaze Tracking},
  booktitle = {Proceedings of the IEEE/CVF International Conference on Computer Vision Workshops (ICCVW)},
  pages     = {1140--1148},
  year      = {2019},
  doi       = {10.1109/ICCVW.2019.00145}
}

@article{yang2021behavioral,
  author    = {Xiaozhi Yang and Ian Krajbich},
  title     = {Webcam-Based Online Eye-Tracking for Behavioral Research},
  journal   = {Judgment and Decision Making},
  volume    = {16},
  number    = {6},
  pages     = {1485--1505},
  year      = {2021},
  doi       = {10.1017/S1930297500008512}
}

@article{kaduk2024lab,
  author    = {Tobiasz Kaduk and Caspar Goeke and Holger Finger and Peter K{\"o}nig},
  title     = {Webcam Eye Tracking Close to Laboratory Standards: Comparing a New Webcam-Based System and the {EyeLink} 1000},
  journal   = {Behavior Research Methods},
  volume    = {56},
  number    = {5},
  pages     = {5002--5022},
  year      = {2024},
  doi       = {10.3758/s13428-023-02237-8}
}

@article{valtakari2024field,
  author    = {Niilo V. Valtakari and Roy S. Hessels and Diederick C. Niehorster and Charlotte Viktorsson and P{\"a}r Nystr{\"o}m and Terje Falck-Ytter and Chantal Kemner and Ignace T. C. Hooge},
  title     = {A Field Test of Computer-Vision-Based Gaze Estimation in Psychology},
  journal   = {Behavior Research Methods},
  volume    = {56},
  number    = {3},
  pages     = {1900--1915},
  year      = {2024},
  doi       = {10.3758/s13428-023-02125-1}
}

@article{vandercruyssen2024validation,
  author    = {Ine Van der Cruyssen and Gershon Ben-Shakhar and Yoni Pertzov and Nitzan Guy and Quinn Cabooter and Lukas J. Gunschera and Bruno Verschuere},
  title     = {The Validation of Online Webcam-Based Eye-Tracking: The Replication of the Cascade Effect, the Novelty Preference, and the Visual World Paradigm},
  journal   = {Behavior Research Methods},
  volume    = {56},
  number    = {5},
  pages     = {4836--4849},
  year      = {2024},
  doi       = {10.3758/s13428-023-02221-2}
}

@article{lei2024endtoend,
  author    = {Yaxiong Lei and Shijing He and Mohamed Khamis and Juan Ye},
  title     = {An End-to-End Review of Gaze Estimation and Its Interactive Applications on Handheld Mobile Devices},
  journal   = {ACM Computing Surveys},
  volume    = {56},
  number    = {2},
  pages     = {1--38},
  year      = {2024},
  doi       = {10.1145/3606947}
}

@article{patterson2025scoping,
  author    = {Allie Spencer Patterson and Christopher Nicklin and Joseph P. Vitta},
  title     = {Methodological Recommendations for Webcam-Based Eye Tracking: A Scoping Review},
  journal   = {Research Methods in Applied Linguistics},
  volume    = {4},
  number    = {3},
  pages     = {100244},
  year      = {2025},
  doi       = {10.1016/j.rmal.2025.100244}
}

\end{document}